\documentclass[twoside,11pt]{article}

\usepackage[preprint]{jmlr2e}
\usepackage[utf8]{inputenc} 
\usepackage[T1]{fontenc}    
\usepackage{hyperref}       
\usepackage{url}   
\usepackage[ruled,longend,linesnumbered]{algorithm2e}
\usepackage{float}
\usepackage{multirow}
\usepackage{booktabs}       
\usepackage{amsfonts}    \usepackage{rotating}
\usepackage{nicefrac}       
\usepackage{microtype}  
\usepackage{graphicx}
\usepackage{tabularx}
\usepackage{subfigure}
\usepackage[dvipsnames]{xcolor}
\usepackage{graphicx}
\usepackage{amsmath}
\usepackage{xcolor}
\usepackage{colortbl}
\usepackage{enumitem}
\usepackage{caption}
\usepackage{listings}
\usepackage{appendix}
\usepackage{chngcntr}
\usepackage{minitoc}

\newcolumntype{C}{>{\centering\arraybackslash}X}

\hypersetup{
    colorlinks = true,
    allcolors = {Tan},
    linkbordercolor = {white},
}

\ShortHeadings{Multitracer Lesion Segmentation in Whole-Body PET/CT}{Qiaoyi Xue et al.}
\firstpageno{1}

\begin{document}

\doparttoc 
\faketableofcontents 

\title{Automated Lesion Segmentation in Whole-Body PET/CT in a multitracer setting}

\author{\name Qiaoyi~Xue * \thanks{These authors contributed to the work equally and should be regarded as co-first authors.} \email 
    qiaoyi.xue@cri-united-imaging.com \\
    \addr Shanghai United Imaging Healthcare Advanced Technology \\ Research Institute Co., Ltd. \\
    Shanghai 201807, China
    \AND
    \name Youdan~Feng * \email youdan.feng@cri-united-imaging.com \\
    \addr Shanghai United Imaging Healthcare Advanced Technology \\ Research Institute Co., Ltd. \\
    Shanghai 201807, China
    \AND
    \name Jiayi~Liu * \email jiayi.liu01@cri-united-imaging.com \\
    \addr Shanghai United Imaging Healthcare Advanced Technology \\ Research Institute Co., Ltd. \\
    Shanghai 201807, China
    \AND
    \name Tianming~Xu \email cecilia\textunderscore xtm@sjtu.edu.cn  \\
    \addr Global Institute of Future Technology \\ Shanghai Jiao Tong University \\
    Shanghai 200240, China
    \AND
    \name Kaixin~Shen \email 812852899@sjtu.edu.cn \\
    \addr Global Institute of Future Technology \\ Shanghai Jiao Tong University \\
    Shanghai 200240, China
    \AND
    \name Chuyun~Shen \email 
    cyshen@stu.ecnu.edu.cn \\
    \addr School of Computer Science and Technology \\
    East China Normal University\\
    Shanghai 200062, China
    \AND
    \name Yuhang~Shi \email yuhang.shi@cri-united-imaging.com \\
    \addr Shanghai United Imaging Healthcare Advanced Technology \\ Research Institute Co., Ltd. \\
    Shanghai 201807, China
}

\maketitle

\begin{abstract}
This study explores a workflow for automated segmentation of lesions in FDG and PSMA PET/CT images. Due to the substantial differences in image characteristics between FDG and PSMA, specialized preprocessing steps are required. Utilizing YOLOv8 for data classification, the FDG and PSMA images are preprocessed separately before feeding them into the segmentation models, aiming to improve lesion segmentation accuracy. The study focuses on evaluating the performance of automated segmentation workflow for multitracer PET images. The findings are expected to provide critical insights for enhancing diagnostic workflows and patient-specific treatment plans.
Our code will be open-sourced and available at \href{https://github.com/jiayiliu-pku/AP2024}{https://github.com/jiayiliu-pku/AP2024}.

\end{abstract}

\section{Introduction}

The increasing global cancer incidence demands advanced diagnostic and therapeutic technologies to enhance precision and personalization in cancer management. Molecular theranostics, which integrates diagnostic imaging with targeted therapy, exemplifies this trend by offering personalized treatment opportunities with unprecedented accuracy. Positron emission tomography (PET) combined with computed tomography (CT) plays a pivotal role in oncological diagnostics, utilizing radiotracers such as Fluorodeoxyglucose (FDG) and prostate-specific membrane antigen (PSMA) to effectively detect and manage various cancers. FDG is particularly effective in highlighting metabolically active cancer cells, facilitating the evaluation of multiple cancer types~\citep{ref-1}. PSMA, highly expressed in prostate cancer cells, is essential for diagnosing and treating prostate cancer, serving as a valuable target for both imaging and therapeutic interventions~\citep{ref-2,ref-3}.

For FDG PET/CT scans, the adoption of the deep learning methods improve lesion segmentation accuracy and overcomes the challenges associated with differentiating pathological changes from physiological uptake in organs like the liver and brain~\citep{ref-4}. Advances in multi-label segmentation techniques enable simultaneous delineation of lesions and high-uptake organs, further improving segmentation accuracy~\citep{ref-5,ref-6}. As PSMA PET imaging has become increasingly vital for early detection of lymph node metastases and monitoring treatment responses, recent research also shows the superiority of using the deep learning methods in segmenting lesion of the PSMA PET images ~\citep{ref-7,ref-8}.

However, a significant challenge lies in the differences between FDG and PSMA PET images, which necessitate specific and targeted preprocessing steps to handle their unique properties. This study aims to develop a lesion segmentation workflow that can effectively manage both FDG and PSMA PET/CT images. Specifically, the study utilizes YOLOv8 to classify FDG and PSMA data and subsequently applies tailored preprocessing techniques before inputting the classified data into dedicated segmentation models for each tracer ~\citep{yolov8}. By evaluating the impact of organ-specific labeling and preprocessing strategies on model performance, this research seeks to optimize PET/CT imaging for broader oncological applications, particularly for individualized prostate cancer interventions.

\section{Methods}
\begin{figure}[htb!]
    \centering
    \includegraphics[width=0.9\linewidth]{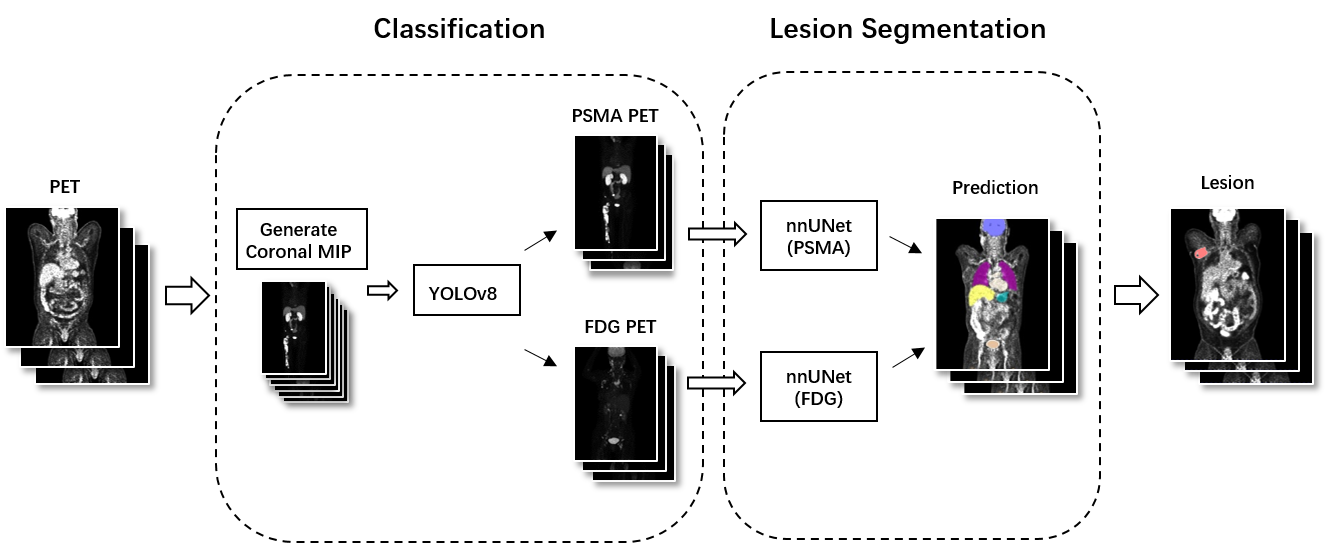}
    \caption{The workflow of automated lesion segmentation of FDG and PSMA PET images. }
    \label{fig:fig1}
\end{figure}

The automated lesion segmentation process for FDG and PSMA PET images consists of two steps. First, a classification model was trained for distinguishing FDG-PET and PSMA-PET medical images. Second, two 3D Unets were trained independently with FDG or PSMA data for the organ and lesion segmentation (shown in Fig.\ref{fig:fig1}). 

\subsection{Data and preprocessing}

\subsubsection{Datasets for lesion segmentation}
The training of the FDG lesion segmentation models was conducted using whole-body FDG PET/CT data from a cohort of 900 patients, encompassing 1014 studies supplied by the AutoPET challenge III in 2024. The challenge consists of patients with malignant melanoma, lymphoma, lung cancer and negative control patients. The data was split into a training set of 811 cases and a testing set of 203 cases. For the PSMA model, 600 PSMA -PET/CT data supplied by the AutoPET challenge III was split into a training set of 500 cases and a testing set of 100 cases. Lesion numbers and patient meta info were taken into consideration to ensure that both the training and testing subsets exhibited equitable distributions of lesion counts. 

\begin{figure}[htb!]
    \centering
    \includegraphics[width=0.6\linewidth]{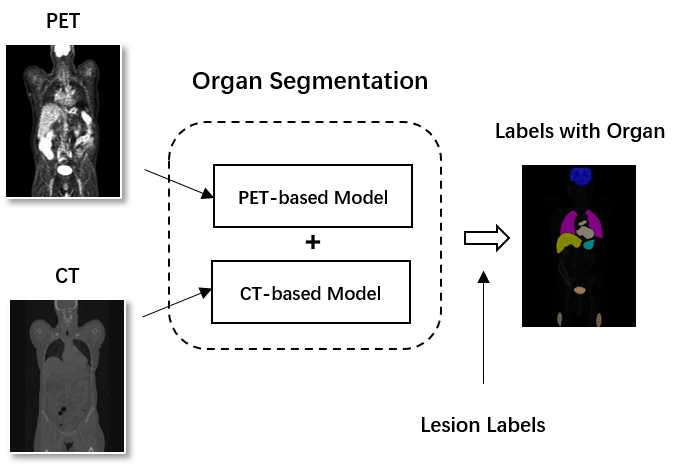}
    \caption{Organ labelling pipeline.}
    \label{fig:fig2}
\end{figure}

In the label preprocessing phase, both CT and PET images were concurrently utilized for organ segmentation. The segmentation of bone structures was achieved using open-source framework Totalsegmentator. The segmentation of high-uptake organs was conducted on PET images using an in-house developed model based on nnU-Net. This approach was meticulously selected to mitigate the potential for mismatch between PET and CT data. Such discrepancies are often attributable to the distinct respiratory phases of abdominal organs during PET/CT scanning; specifically, CT scans are typically acquired during breath-hold periods, whereas PET scans are acquired over several minutes, capturing an average representation of the free-breathing state. This phenomenon is particularly pronounced in the case of the liver and lungs. The bone and organ segmentation labels (liver, kidneys, urinary bladder, spleen, lung, brain, heart, femur, stomach and prostate) were subsequently integrated with lesion labels, which were provided as part of the AutoPET challenge dataset (shown in Fig.\ref{fig:fig2}).  

The data preprocessing procedures were integrated within the nnU-Net pipeline. In brief, the images underwent resampling to achieve uniform voxel spacing, followed by intensity normalization through the computation of the z-score. This standardized preprocessing ensures that the data is primed for robust and consistent analysis within the framework of the nnU-Net neural network.

\subsubsection{Datasets for image classification}
It is observed that PSMA-PET shows higher uptake than FDG in submandibular glands, kidneys, liver, spleen and bladder. Based on the observation, two steps were conducted to classify FDG-PET and PSMA-PET images. The maximum-intensity projection (MIP) images were generated by projecting the voxel with the highest FDG uptake value on coronal view throughout the volume onto a 2D image. Besides the Data supplied by the AutoPET challenge III, additional in-house FDG and PSMA PET MIP images are collected. The training dataset comprised of 1378 FDG and 1345 MIP PSMA images, while the testing dataset include 474 FDG and 539 PSMA MIP images. Images were preprocessed by resizing to 640x640 pixels and normalizing pixel values.

\subsection{Model architecture and training}
\subsubsection{YOLO model}
The YOLOv8 architecture was adapted for classification of PSMA-PET and FDG-PET. The model was trained with hyperparameters optimized: initial learning rate set to 0.0001 and batch size to 16. Training spanned 200 epochs, incorporating data augmentation techniques like flipping to enhance model robustness and mitigate overfitting.  

\subsubsection{nnU-Net model}
The models were trained based on the nnU-Net framework to segment multiple organs and lesions ~\citep{nnunet}. 3D nnU-Net was used with the ResNet18 backbone structure. The input patch size of the 3D U-Net was set to 160x160x160. The loss function is set to a combination of the Dice loss and focal loss to combat overfitting. 
\begin{itemize}
\item \textbf{Dice Loss}:
The Dice Loss is based on the Dice coefficient, which is a measure of overlap between two sets. The formula for Dice Loss is:

	\begin{equation}
	\text{Dice Loss} = 1 - \frac{2 \sum_{i} p_i g_i}{\sum_{i} p_i^2 + \sum_{i} g_i^2}
	\end{equation}
 
where $p_{i}$ is the predicted probability for pixel $i$, $g_{i}$ is the ground truth label for pixel $i$. A higher Dice coefficient indicates a greater overlap between the predicted segmentation and the ground truth, reflecting a more accurate segmentation result.

\item \textbf{Focal Loss}:
The Focal Loss is designed to address class imbalance by down-weighting the loss assigned to well-classified examples. The Focal Loss is defined as:
\begin{equation}
\text{Focal Loss}(p_{t}) = -\alpha_t (1 - p_{t})^\gamma \log(p_{t})
\end{equation}
where $p_{t}$ is the predicted probability of the true class. $\alpha_{t}$ is a weighting factor for class imbalance. $\gamma$ is the focusing parameter that controls the rate at which easy examples are down-weighted.
\end{itemize}

The models are trained with 1000 epochs and a batch size of 4, using the SGD optimizer and an initial learning rate of 0.01. This model was then formatted in docker and submitted to the challenge portal for testing and benchmarking.

The evaluation of model performance is conducted using metrics such as the Dice score, false positive volume (FPvol) and false negative volume (FNvol), which provide a comprehensive assessment of the segmentation methodologies.

\section{Results and Discussion}

The PET model achieved a classification accuracy of 99.85\% which showed superior performance in differentiating FDG and PSMA PET images.  

The evaluation of our lesion segmentation models for FDG and PSMA PET/CT images produced the following outcomes (shown in Table.\ref{tab:tab1}): the Dice coefficients were 0.8408 for FDG and 0.7385 for PSMA. False Positive volumes (FPvol) were 1.7979 for FDG and 9.3574 for PSMA, while False Negative volumes (FNvol) were 2.3625 for FDG and 5.0745 for PSMA. These results indicate differences in segmentation performance between the two imaging modalities.

\begin{table}[]
\caption{Performance of lesion segmentation models for PSMA and FDG PET images.}
\centering
\resizebox{0.6\textwidth}{!}{%
\begin{tabular}{@{}l|c|c|c@{}}
\toprule
\textbf{Method}                 & \textbf{Dice}    & \textbf{FPvol}    & \textbf{FNvol}   \\ 
\midrule
FDG nnU-Net       &  0.8408     & 1.7979    & 2.3625      \\
PSMA nnU-Net       & 0.7385     & 9.3574    & 5.0745      \\
\bottomrule
\end{tabular}}
\label{tab:tab1}
\end{table}

\section{Conclusion}

In conclusion, this study demonstrated the feasibility of the proposed lesion segmentation workflow for both FDG and PSMA PET/CT images. YOLOv8 demonstrated its superior performance in classifying the PSMA and FDG PET image which allows for using tailored preprocessing techniques in segmenting the lesion in PET image with different tracers. 

\clearpage
\newpage

\bibliography{main}

\end{document}